\documentclass[]{spie}  
\usepackage{amsmath,graphicx}
\usepackage{hyperref}
\usepackage{enumitem}
\usepackage{url}
\setlist{nosep, leftmargin=14pt}

\usepackage{mwe} 

\title{Evaluating Detection Thresholds: The Impact of False Positives and Negatives on Super-Resolution Ultrasound Localization Microscopy}
\author[a]{Sepideh K. Gharamaleki}

\author[b, c]{Brandon Helfield}
\author[a]{Hassan Rivaz}
\affil[a]{Concordia University, Department of Electrical and Computer Engineering, Montreal, QC, H3G 2W1, Canada}
\affil[b]{Concordia University, Department of Physics, Montreal, QC, H4B 1R6, Canada}
\affil[c]{Concordia University, Department of Biology, Montreal, QC, H4B 1R6, Canada}
\begin{document}
%
\maketitle
\begin{abstract}

Super-resolution ultrasound imaging with ultrasound localization microscopy (ULM) offers a high-resolution view of microvascular structures. Yet, ULM image quality heavily relies on precise microbubble (MB) detection. Despite the crucial role of localization algorithms, there has been limited focus on the practical pitfalls in MB detection tasks such as setting the detection threshold. This study examines how False Positives (FPs) and False Negatives (FNs) affect ULM image quality by systematically adding controlled detection errors to simulated data. Results indicate that while both FP and FN rates impact Peak Signal-to-Noise Ratio (PSNR) similarly, increasing FP rates from 0\% to 20\% decreases Structural Similarity Index (SSIM) by 7\%, whereas same FN rates cause a greater drop of around 45\%. Moreover, dense MB regions are more resilient to detection errors, while sparse regions show high sensitivity, showcasing the need for robust MB detection frameworks to enhance super-resolution imaging. The code is available at \url{http://code.sonography.ai/} and \href{https://github.com/IMPACT-L/ULMThresholdAnalysis}{https://github.com/IMPACT-L/ULMThresholdAnalysis}.

\end{abstract}
\keywords{Super-resolution ultrasound, Ultrasound Localization Microscopy, Microbubbles, Sensitivity analysis.} 

\section{Introduction}
\label{sec:intro}
Ultrasound imaging has long been a cornerstone of clinical diagnostics, valued for its versatility and accessibility. However, traditional US imaging faces an inherent limitation in spatial resolution, which is fundamentally constrained by the physics of acoustic waves \cite{Swanevelder2011, Goudarzi2023_UnifyingUltrasoundBeamforming}. The resolution limit is approximately half the wavelength of the employed acoustic wave, creating a barrier to visualizing fine anatomical structures.
Super-Resolution UltraSound (SR-US) represents a leap in vascular imaging resolution by providing the means to reconstruct the vascular network and achieve a resolution close to one-tenth of the imaging wavelength of conventional diffraction-limited ultrasound \cite{Hess2006}.

Ultrasound Localization Microscopy (ULM) currently represents the only method that combines cost-effectiveness, non-invasiveness, and non-ionizing properties for comprehensive in vivo imaging of microvasculature. Through SR-US, low-resolution ultrasound images are transformed into high-resolution images, enabling improved diagnostic accuracy. Studying the vascular system can aid in the diagnosis and treatment of cardiovascular \cite{Zhang2023}, neurodegenerative disease \cite{Renaudin2022} and cancer \cite{Porte2024}. This wide-ranging applicability underscores ULM's potential to revolutionize our understanding of microvascular systems in research and clinical settings.

Christensen-Jeffries et al. \cite{ChristensenJeffries2020} made a significant breakthrough in ultrasound imaging with their pioneering technique of tracking individual microbubbles (MBs) within the vascular system.  Their work demonstrated that both acoustic super-resolution and sub-resolution velocity imaging can be achieved using standard image data from clinical ultrasound systems by sub-pixel localization and tracking of MBs. MBs are gas-filled, micron-scale contrast agents encapsulated by a stabilizing shell (e.g., lipid or polymer) that undergo nonlinear oscillations when excited by an ultrasound field, producing strong, distinct acoustic responses that enable their detection and localization.

Despite its promise, SR-US faces significant hurdles in its transition to clinical applications, namely the insufficient detection of MBs within a specified accumulation period, the time required for data collection and the imaging depth \cite{Hingot2019}. These limitations constrain the technique's ability to provide high-resolution, real-time imaging in clinical settings, in the sense that optimizing any one of these factors often necessitates compromises in the others. These challenges collectively highlight the complexities involved in advanced imaging techniques, particularly in medical and diagnostic applications. 
In this context, optimization-based algorithms have gained prominence as a technique aimed at enhancing the resolution of low-resolution ultrasound images by optimizing a cost function or likelihood based on the signal model for MB localization. \cite{ashikuzzaman2022,dencks2020,diamantis2018,diamantis2019}. However, these methods depend on the assumption of isolated scatterers within the region of interest \cite{Heiles2022} which is no longer correct as MB trajectories intersect.

To address the limitations of conventional methods, researchers have turned to learning-based approaches. In recent years, deep learning algorithms have emerged as the dominant paradigm in ultrasound image analysis ~\cite{Goudarzi2023_SegmentationArmUltrasound_TBME} and reconstruction~\cite{Goudarzi2022_InverseUltrasoundBeamforming_TUFFC}, consistently achieving state-of-the-art performance across various benchmarks and applications. While the introduced networks often share certain components or characteristics, they can differ significantly in their overall structure and specific implementations \cite{vanSloun2019,Milecki2021,Chen2022,gharamaleki2022,gharamaleki2023, Gharamaleki2025_DeformableDETR_ULM}.


These methods mark significant progress in deep learning-based localization for ultrasound imaging; however, they rely on numerous parameters that often need to be manually set, either through informed estimation or trial and error. As localization solutions vary significantly depending on training and simulation parameters, the importance of proper parameter selection becomes even more critical. However, there has been little to no systematic analysis of this selection process, and our aim here is to address this gap.

Choosing the correct threshold for detection scores can be crucial for balancing precision and recall, which directly affects the performance of the model and, subsequently, the SR map. The optimal threshold depends on several factors, including the imaging parameters, the tolerance for False Positives (FPs) and False Negatives (FNs), and the distribution of MB sizes and locations in the data.

In this paper, we:

1. Investigate the trade-off between sensitivity and specificity, with an emphasis on identifying which factor has a more significant impact on performance.

2. Assess multiple metrics of SR maps to evaluate this trade-off across frames, focusing on different density regions.

3. Explore the influence of varying center frequencies on the relationship between sensitivity and specificity.

\section{Methods}
\label{sec:methods}

The primary aim of this study is to quantitatively assess the effects of increasing FPs and FNs on the quality of SR maps derived from ULM. The experiment utilizes two initial datasets of Ground Truth (GT) locations from the IEEE UltraSR Challenge \cite{Lerendegui2024, Lerendegui2022_ULTRASR_Data}, with different center frequencies. We name these two Simulation 1 for the simulation with a center frequency of 2.841MHz and Simulation 2 for the simulation with a center frequency of 7.24MHz.

\begin{figure}[htb]
    \centering
    (A)\\
    \includegraphics[width=0.141\textwidth,height=0.17\textheight]{ 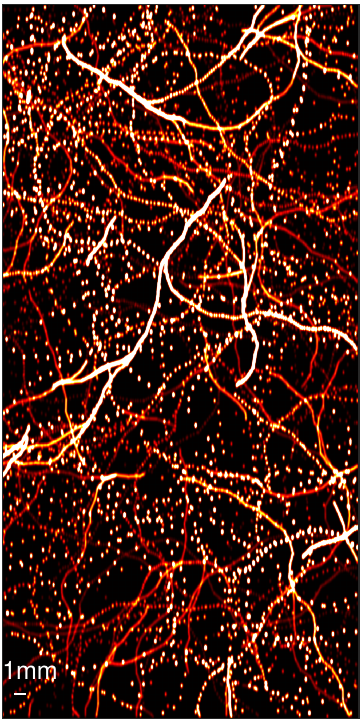}
    \hspace{0.02\textwidth}
    \includegraphics[width=0.141\textwidth,height=0.17\textheight]{ 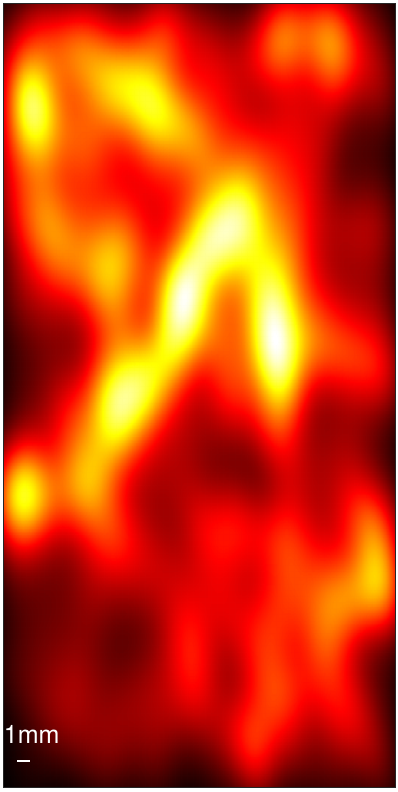} 
    \hspace{0.02\textwidth}
    \includegraphics[width=0.141\textwidth,height=0.17\textheight]{ 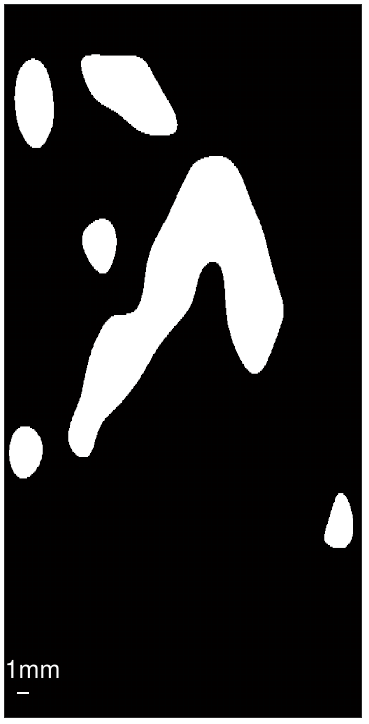}
    \\
    (B)\\
    \includegraphics[width=0.141\textwidth,height=0.17\textheight]{ 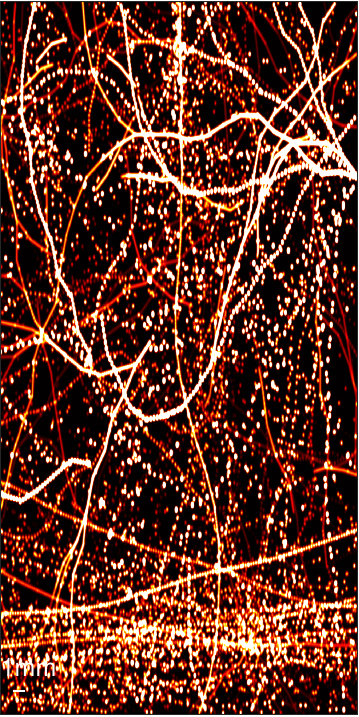} 
    \hspace{0.02\textwidth}
    \includegraphics[width=0.141\textwidth,height=0.17\textheight]{ 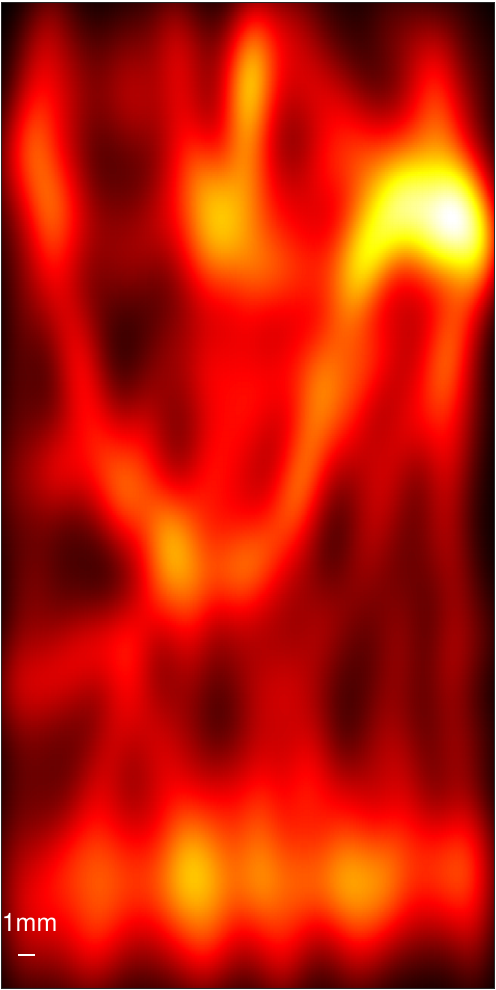} 
    \hspace{0.02\textwidth}
    \includegraphics[width=0.141\textwidth,height=0.17\textheight]{ 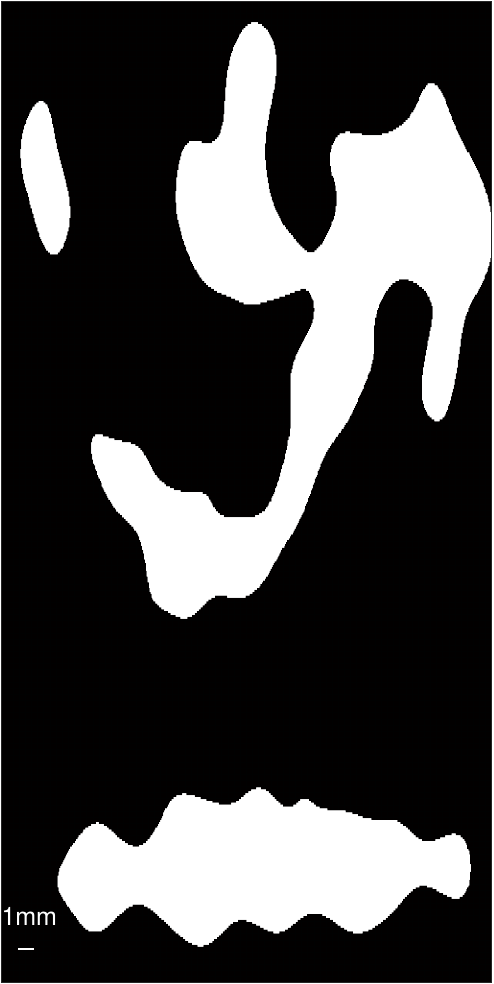}
   
     \begin{tabular}{p{0.14\textwidth}p{0.14\textwidth}p{0.14\textwidth}}
        \centering \makebox[0.14\textwidth]{SR Map} & 
        \centering \makebox[0.14\textwidth]{KDE analysis} & 
        \centering \makebox[0.14\textwidth]{Region Masks}
    \end{tabular}
    \vspace{0.5cm} 
    \caption{left: SR Maps, middle: KDE analysis and right: Region Masks (black: Sparse and white: Dense regions) for (A) Simulation 1 and (B) Simulation 2}
    \label{fig:regions}
\end{figure}

\begin{figure*}[htb]
    \centering
    (A)\\
    \includegraphics[width=0.3\textwidth,height=0.2\textheight]{ 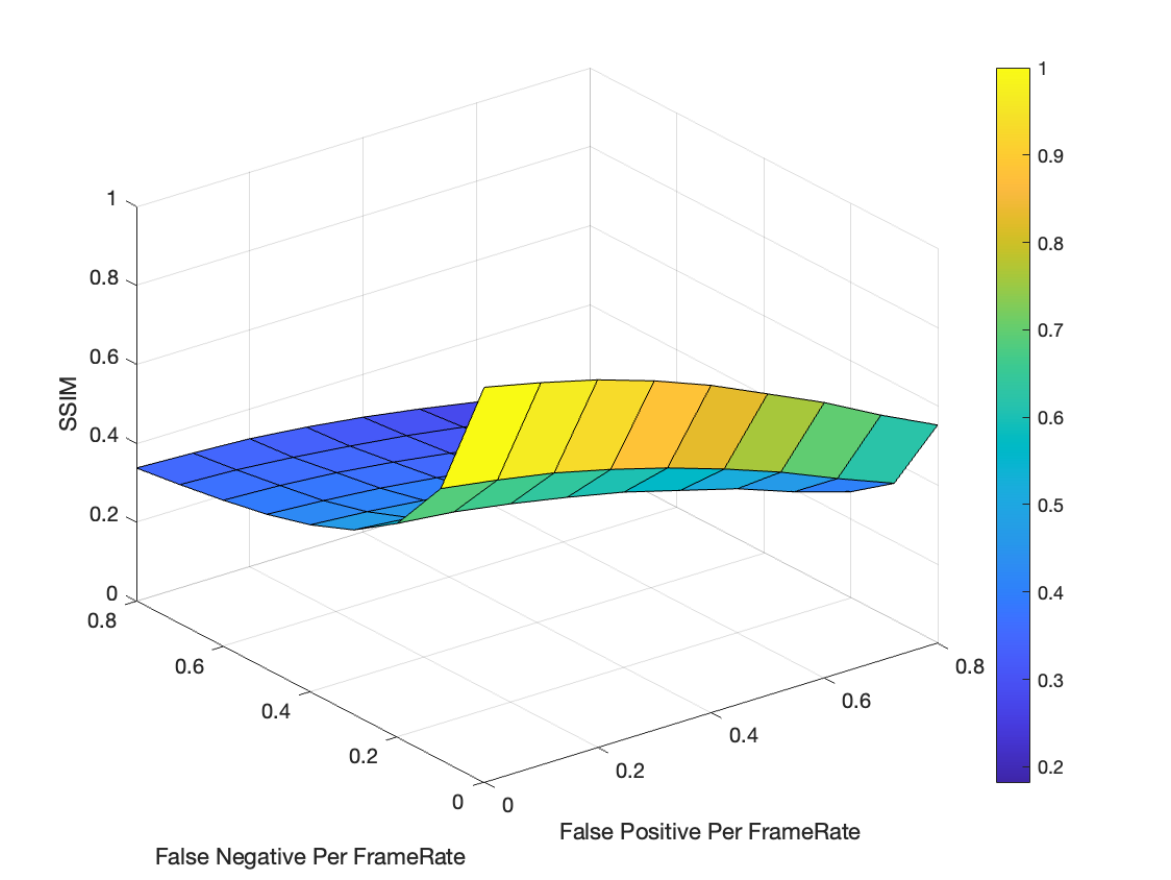}
    \hspace{0.02\textwidth}
    \includegraphics[width=0.3\textwidth,height=0.2\textheight]{ 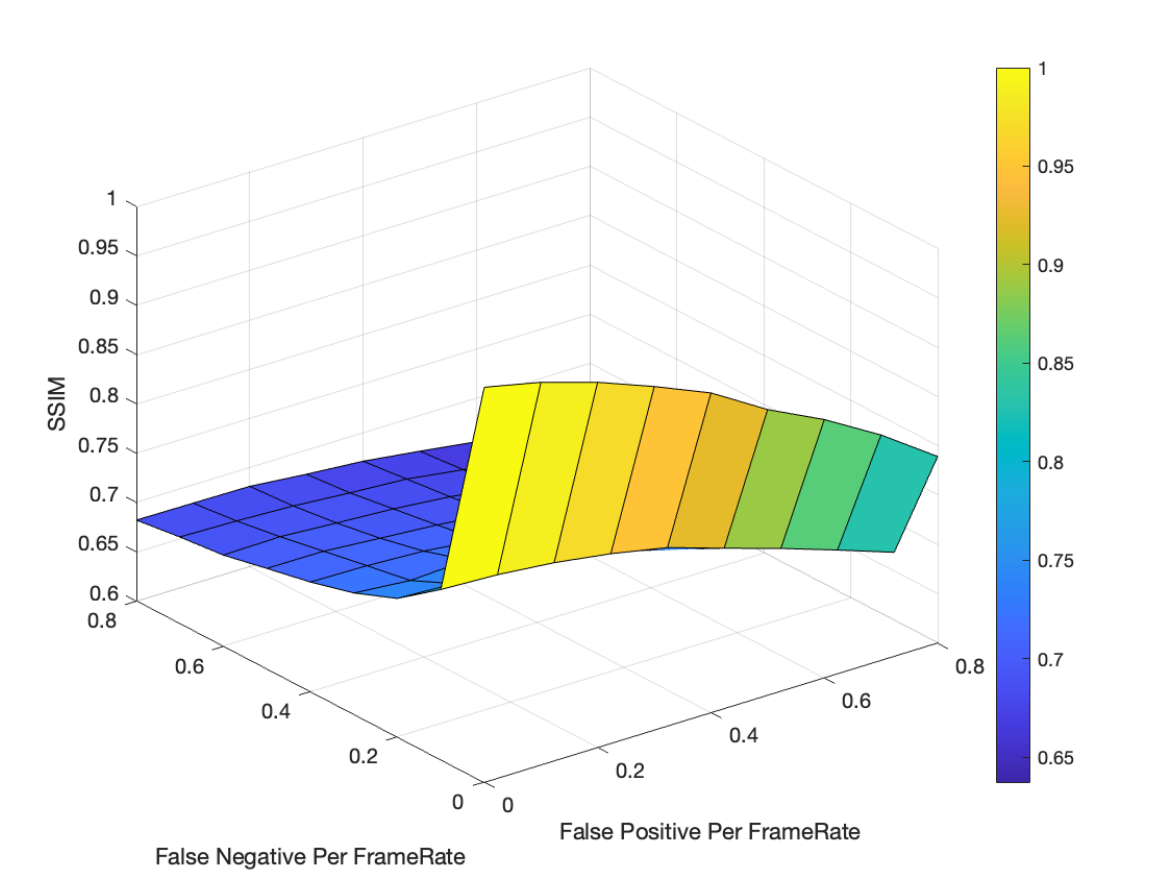} 
    \hspace{0.02\textwidth}
    \includegraphics[width=0.3\textwidth,height=0.2\textheight]{ 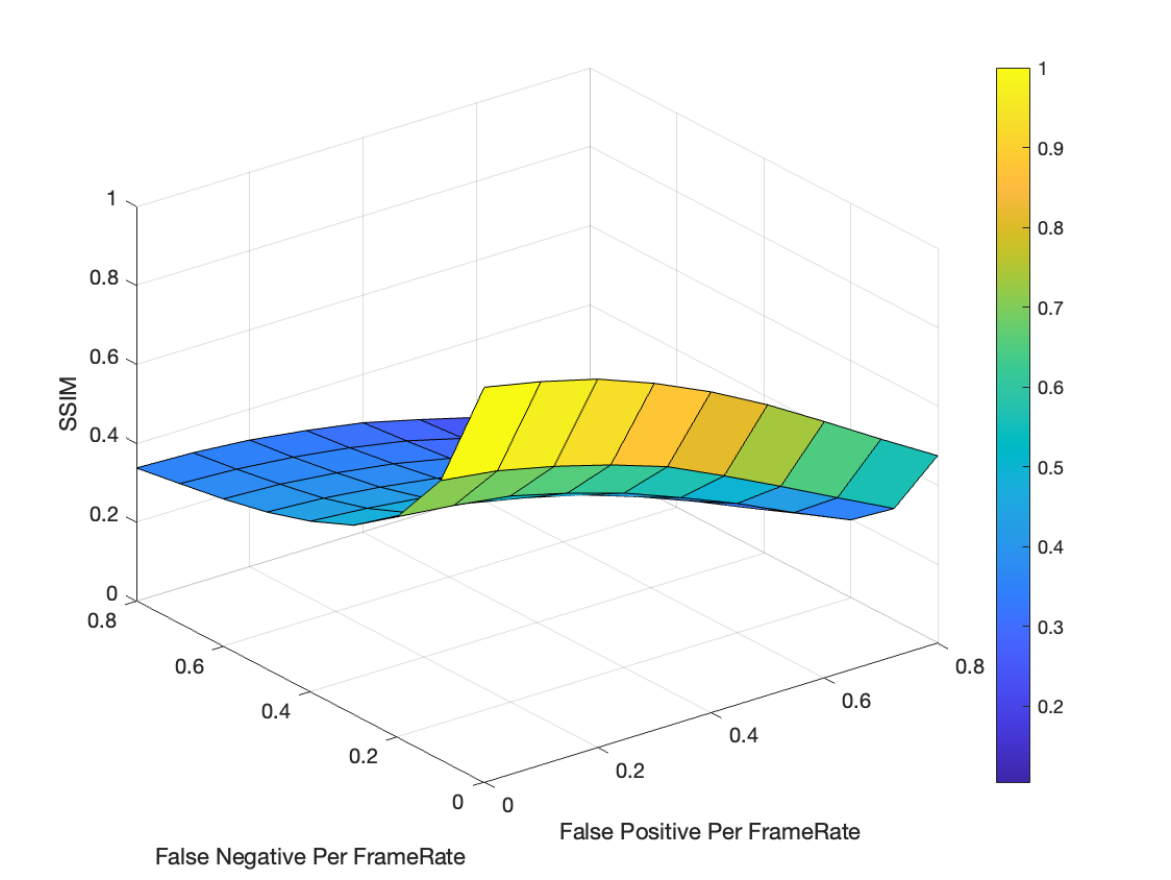}
    \\
    (B)\\
    \includegraphics[width=0.3\textwidth,height=0.2\textheight]{ 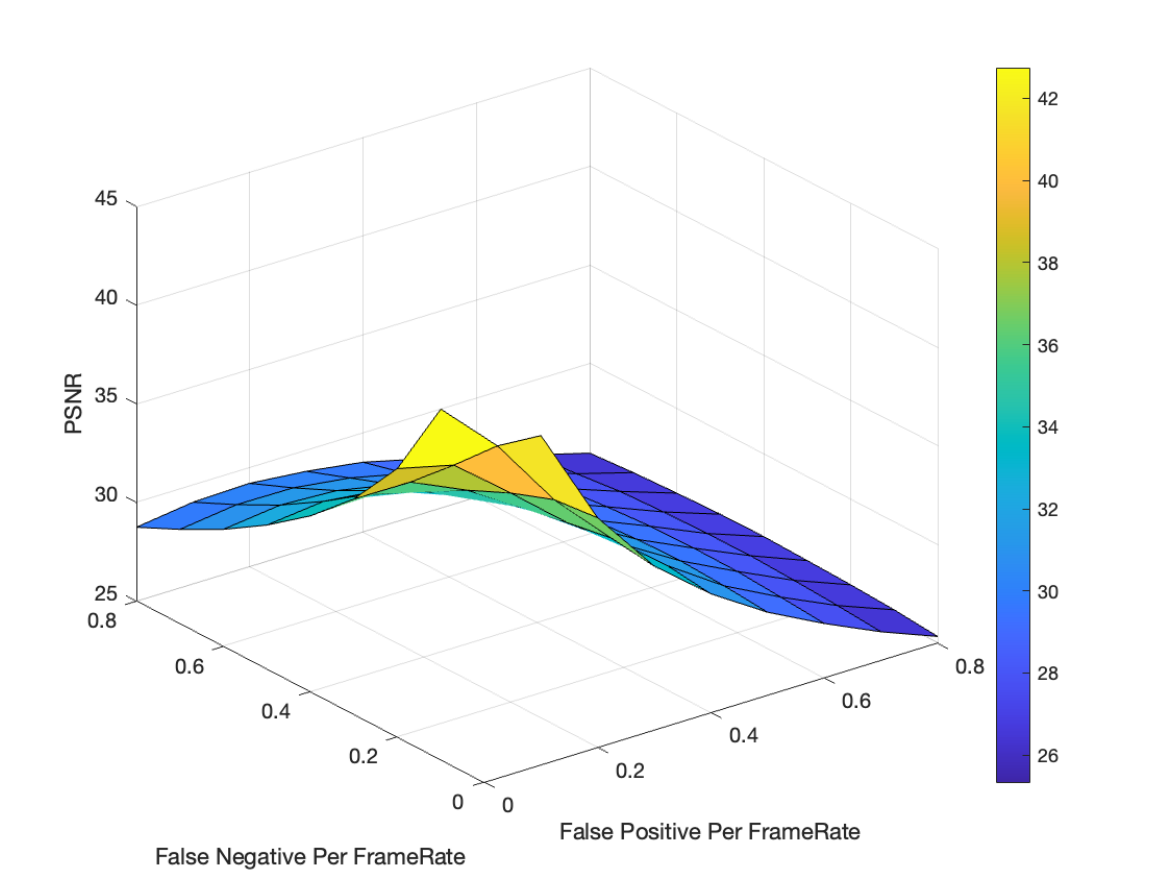} 
    \hspace{0.02\textwidth}
    \includegraphics[width=0.3\textwidth,height=0.2\textheight]{ 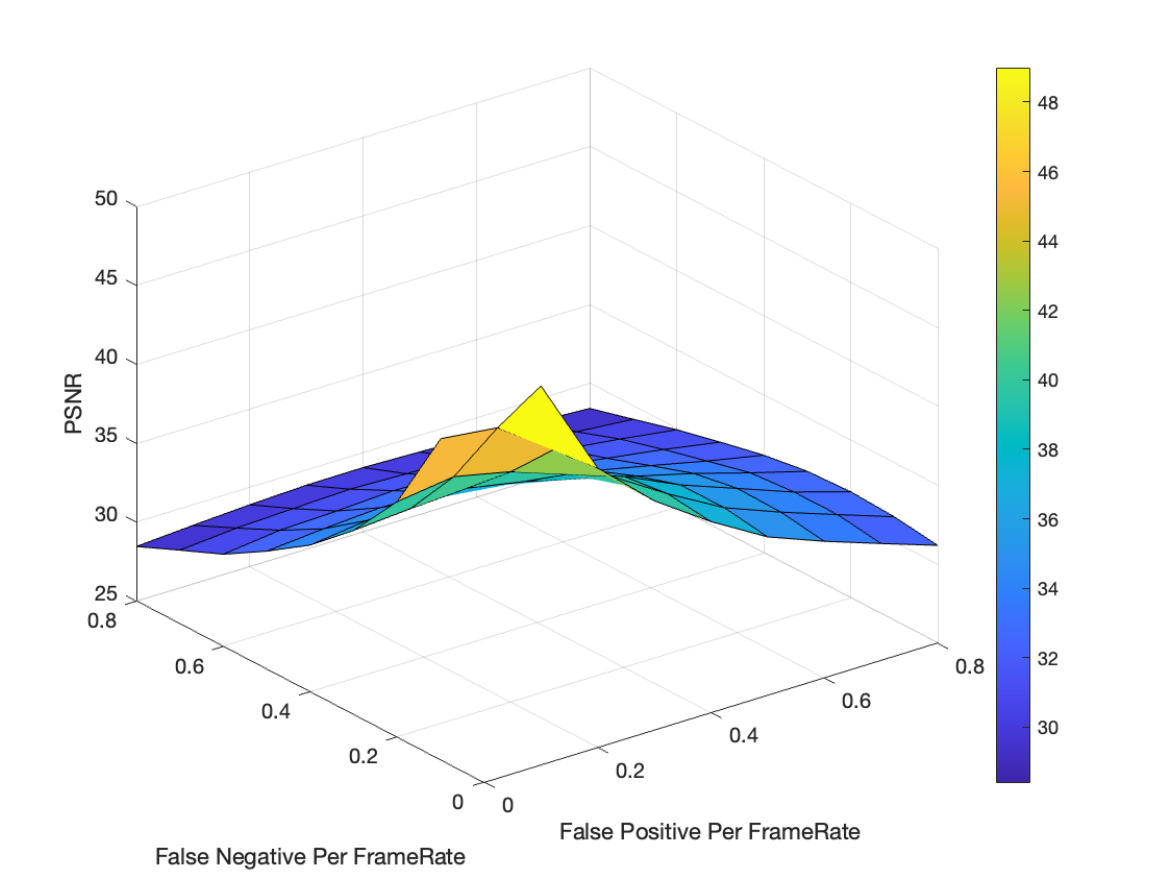} 
    \hspace{0.02\textwidth}
    \includegraphics[width=0.3\textwidth,height=0.2\textheight]{ 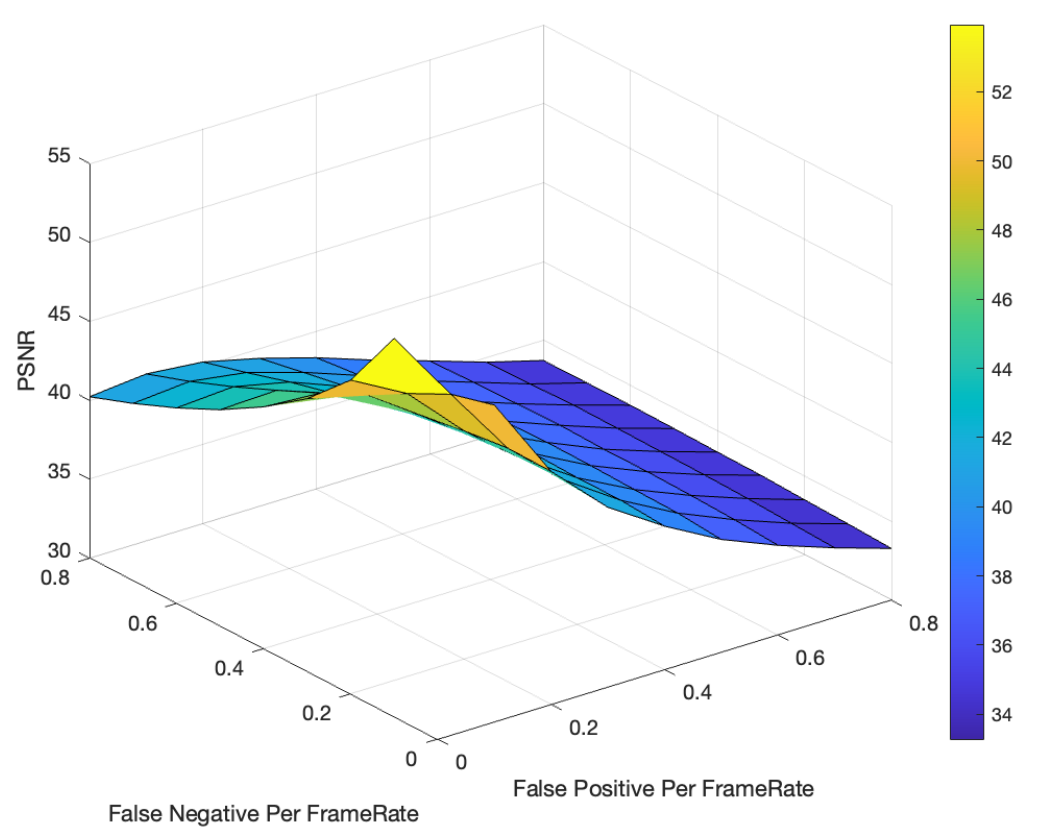}
   
     \begin{tabular}{p{0.3\textwidth}p{0.3\textwidth}p{0.3\textwidth}}
        \centering \makebox[0.3\textwidth]{Sparse regions} & 
        \centering \makebox[0.3\textwidth]{Dense regions} & 
        \centering \makebox[0.3\textwidth]{All regions}
    \end{tabular}
    \vspace{0.5cm} 
    \caption{Result of varying FP and FNs in Sparse, dense and all the regions of image (indicated by columns) for (A) SSIM (B) PSNR of Simulation 1 }
    \label{fig:simu1}
\end{figure*}

\begin{figure*}[htb]
    \centering
    (A)\\
    \includegraphics[width=0.3\textwidth,height=0.2\textheight]{ 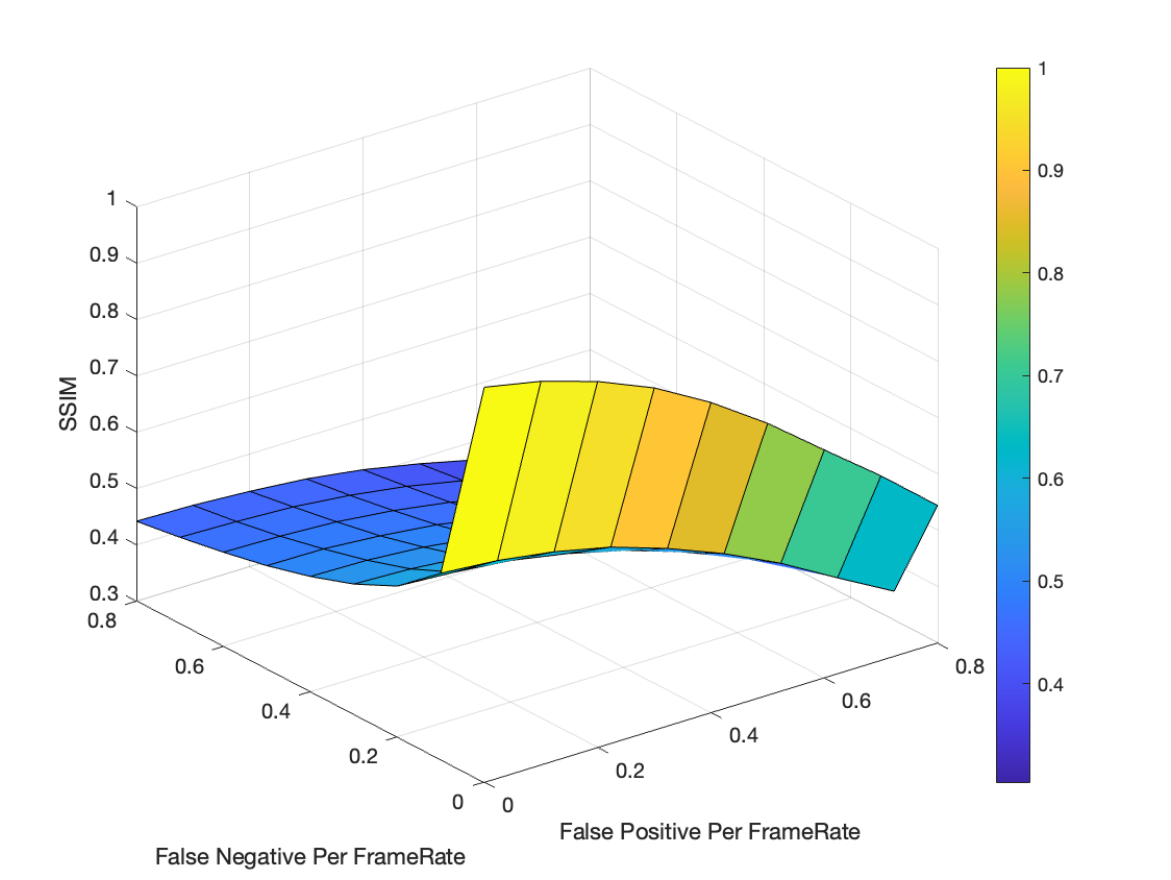}
    \hspace{0.02\textwidth}
    \includegraphics[width=0.3\textwidth,height=0.2\textheight]{ 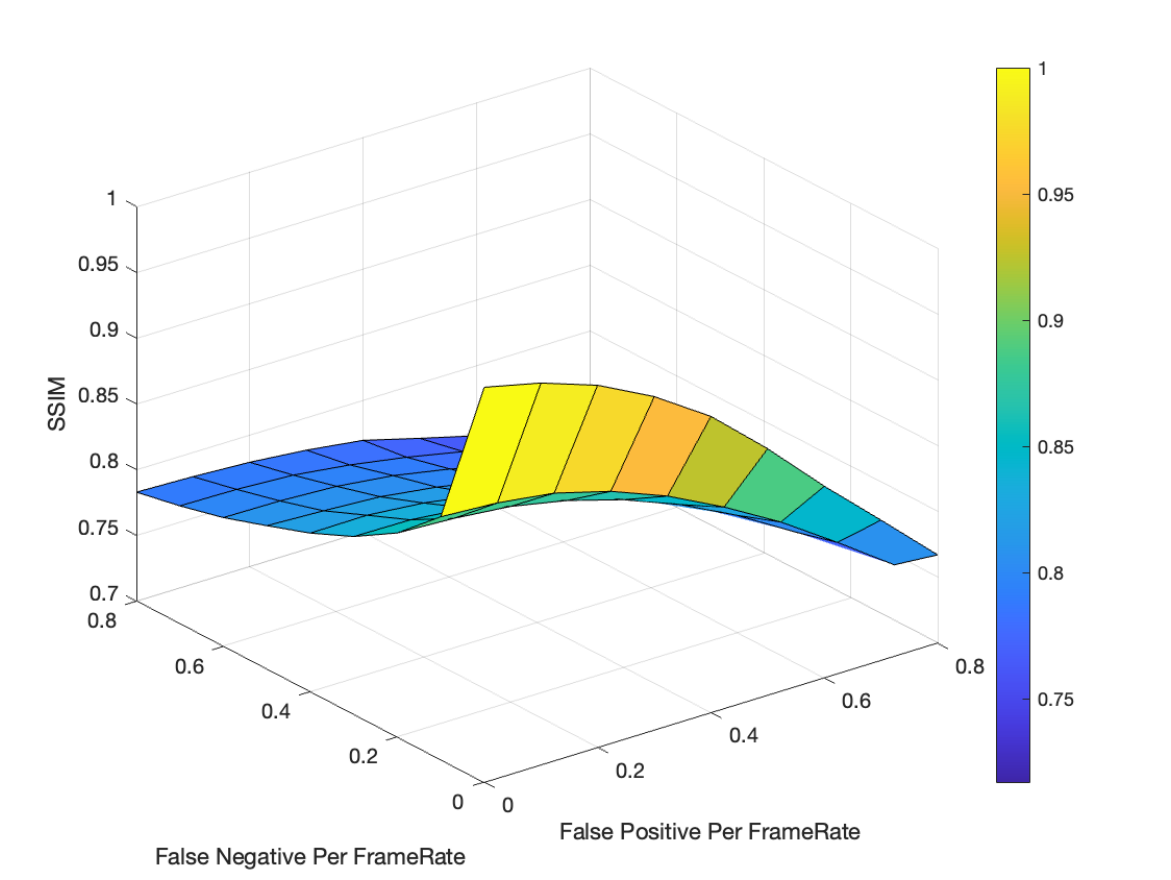} 
    \hspace{0.02\textwidth}
    \includegraphics[width=0.3\textwidth,height=0.2\textheight]{ 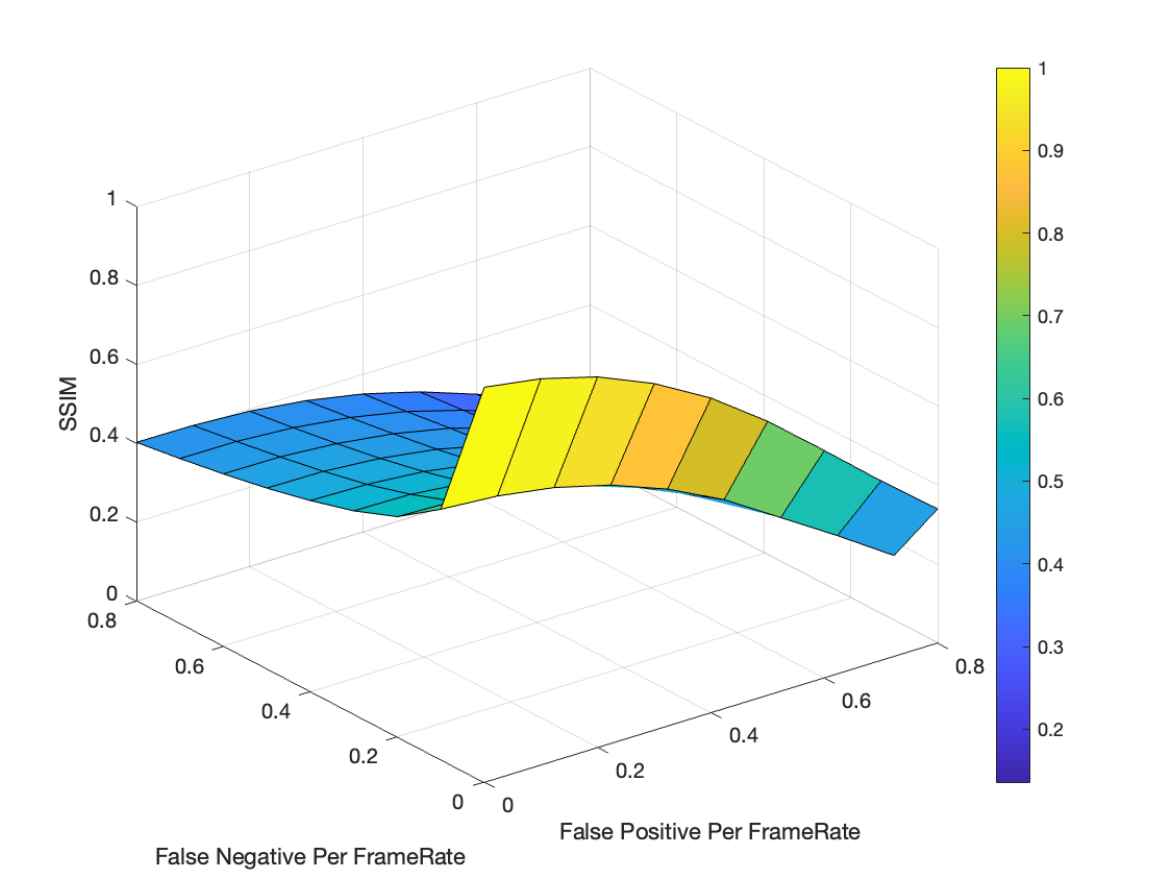}
    \\
    (B)\\
    \includegraphics[width=0.3\textwidth,height=0.2\textheight]{ 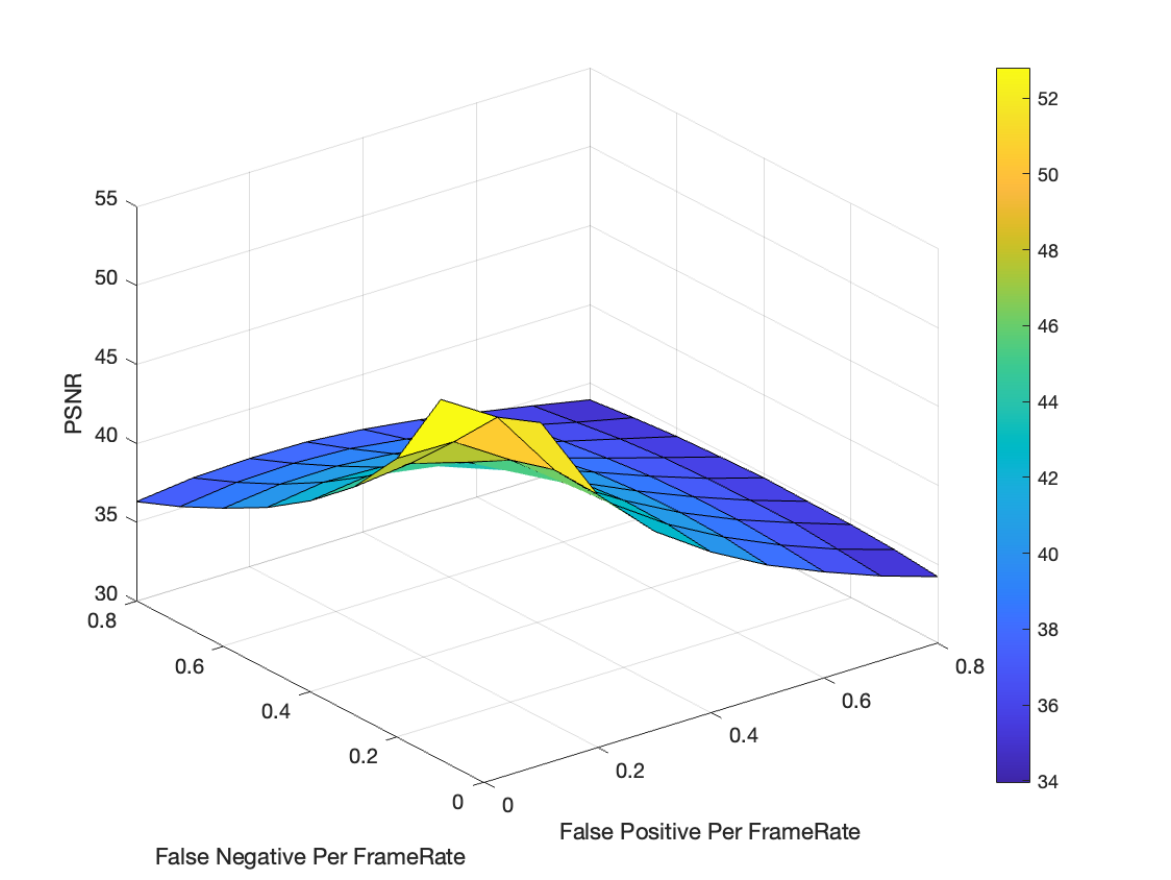} 
    \hspace{0.02\textwidth}
    \includegraphics[width=0.3\textwidth,height=0.2\textheight]{ 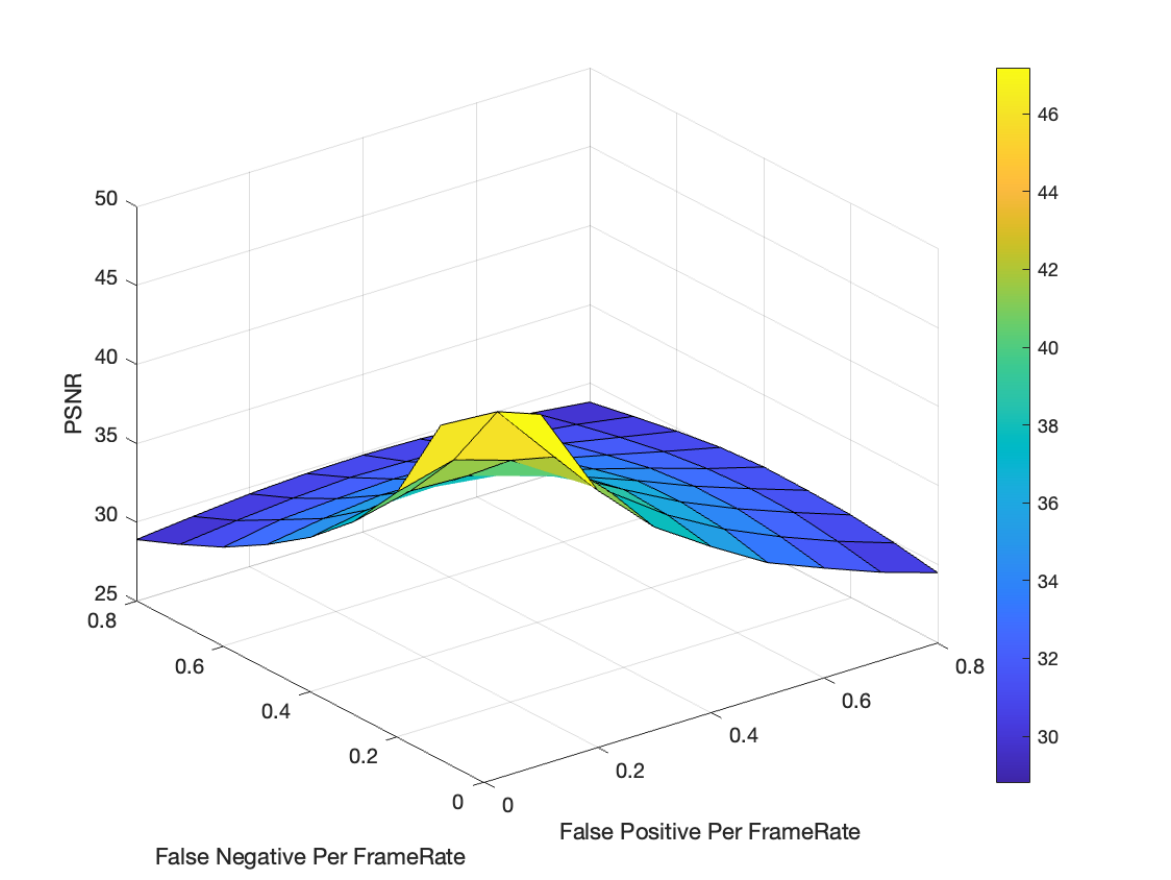} 
    \hspace{0.02\textwidth}
    \includegraphics[width=0.3\textwidth,height=0.2\textheight]{ 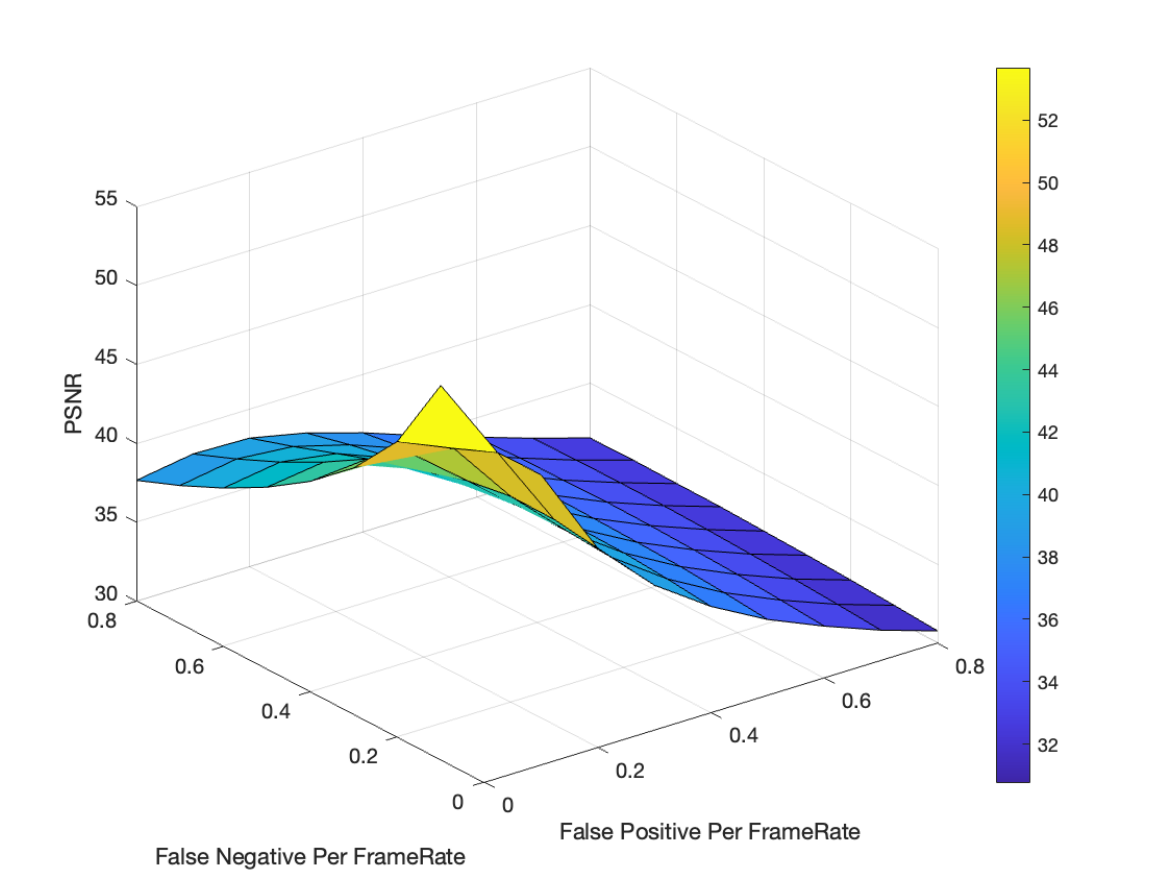}

      \begin{tabular}{p{0.3\textwidth}p{0.3\textwidth}p{0.3\textwidth}}
        \centering \makebox[0.3\textwidth]{Sparse regions} & 
        \centering \makebox[0.3\textwidth]{Dense regions} & 
        \centering \makebox[0.3\textwidth]{All regions}
    \end{tabular}
    \vspace{0.5cm}  
    \caption{Result of varying FP and FNs in Sparse, dense and all the regions of image (indicated by columns) for (A) SSIM and (B) PSNR of Simulation 2.}
    \label{fig:simu2}
\end{figure*}

To evaluate the impact of FPs, we randomly introduce FPs into each frame of the datasets. The number of FPs added is varied across multiple scenarios, allowing for an exploration of a range of error rates and their respective effects on SR map quality. Additionally, certain GT MBs were randomly removed, thereby introducing FNs into the dataset. The proportion of GT locations removed is adjusted to enable a controlled examination of FN impact on the quality of SR maps. To quantify the quality of the SR maps, two key metrics are used: the Structural Similarity Index (SSIM), which measures the similarity between the generated SR map and the GT SR map, to highlight perceptual differences and the Peak Signal-to-Noise Ratio (PSNR), which assesses the quality of reconstruction in the SR maps. SR maps corresponding to each scenario are generated and visualized for qualitative assessment. Furthermore, the computed SSIM and PSNR values are plotted against varying FP counts and FN percentages to elucidate their relationship with SR map quality.

Next, to further understand the effects of MB detection density on the sensitivity to FPs and FNs, the dataset is segmented into different density regions, specifically comparing sparse and dense regions of MBs (Fig. \ref{fig:regions}). 
To analyze the spatial distribution of objects across the 2D plane, a Gaussian Kernel Density Estimate (KDE) was employed. KDE is a non-parametric way that provides a smoothed representation of the distribution of objects based on observed coordinates. The results were visualized as heatmaps in Fig. \ref{fig:regions}, where the color intensity corresponds to the estimated object density at each location.  The application of the KDE allows for a more intuitive visualization of object concentration, especially when compared to direct point-based representations of object locations. 
This smoothed density map was then thresholded to determine the mask for dense and sparse regions.

For each density region, FPs and FNs are artificially injected and the quality metrics are calculated separately for each to determine whether the effects of FPs and FNs vary based on MB density. This comparative analysis aims to provide insights into the relationship between detection density and error sensitivity, enabling a deeper understanding of how varying levels of MB concentration influence the overall quality of SR images.

\section{Results}
\label{sec:results}

SSIM and PSNR metrics for a grid of FPs and FNs with different rates have been calculated and visualized in Fig. \ref{fig:simu1} and \ref{fig:simu2} for Simulation 1 and 2, respectively, with each column showing the results for different regions. 

For both simulations, baseline SSIM values for sparse regions are consistently lower compared to dense regions (Figs. 2 and 3 (A)). The higher baseline metrics in dense regions can be attributed to several factors, including increased signal redundancy, enhanced structural information and higher local contrast. These characteristics lead to greater robustness against errors in dense areas, as reflected in the higher SSIM values observed.

In dense regions, SSIM values remained relatively high, even as FP and FN rates increased with SSIM gradually declining to a minimum of 0.6375 for Simulation 1 and 0.7171 for Simulation 2, as error rates increased. This stability in dense regions, with SSIM values generally above 0.6, indicates a resilience to moderate detection errors, though higher FNs led to a steeper decline in SSIM, compared to FP rates, emphasizing the need to manage FN rates to preserve structural information. 

In sparse regions, SSIM displayed a greater sensitivity to both FP and FN rates, with values quickly decreasing as FPs and FNs increased, reaching as low as 0.2409 and 0.3041 for simulations 1 and 2, respectively. Furthermore, missed detections (high FN rates) disrupt structural similarity more severely, which could be attributed to moderate FP rates appearing to ``fill in'' missing information, thereby contributing to the contextual information, and a lower drop in metrics.

For simulation 2 (Fig. 3 (A)), i.e., higher frequency imaging, even more noticeable sensitivity to FN rates is observed, with around 10\% more drop in SSIM when adding 10\% MBs as FNs to each frame in sparse regions. This indicates that missed detections, in sparse regions at higher frequencies, have a greater impact on structural similarity, making them particularly vulnerable to detection errors. Although after adding the first 10\% of FNs, the gradient of changes in higher and lower frequency remains mostly the same.

PSNR values for sparse regions (Fig. 2 and 3 (B)) remain relatively high even as FP and FN rates increase. At low FP and FN rates, PSNR values are highest, with values above 40 dB for both simulations. As detection errors increase, PSNR declines to approximately 28 dB with high FN and FP rates. The sparse region shows a steeper decline in PSNR as FP increases, which shows the resilience of dense regions to detection errors. This can be attributed to the fact that, in dense regions, the effect of FPs is less pronounced, as their contribution as noise is diminished by the inherent structural redundancy. In other words, for dense regions, FP instances can be partially covered by TP, whereas in sparse ones they stand as isolated bright points.  

\section{Discussion}

SSIM outperforms PSNR in evaluating super-resolution ultrasound images by considering contrast, structural integrity and aligning better with human visual perception. Unlike PSNR, which measures only pixel-wise error, SSIM reflects visual and structural quality, crucial in ULM where preserving details matters. Notably, increasing FP rates from 0 to 20\% reduces SSIM by 7\% in all regions, while a similar FN increase causes a larger SSIM decline of 45\% in the same regions, though both rates impact Peak PSNR similarly.


Data at higher frequencies, in general, exhibited greater sensitivity. When comparing the low and high center frequency data, the results reveal a trade-off: higher frequencies enhance both SSIM and PSNR metrics under optimal detection settings; however, high-frequency ULM images are likely more vulnerable to quality degradation from FNs. Therefore, to leverage the benefits of higher frequencies, the detection thresholds must be finely tuned to mitigate FNs while balancing FP rates.

Overall, these findings emphasize the need for region-specific thresholds to optimize image quality, with dense regions benefiting from higher baseline metrics while also showing greater adaptability to moderate error rates.

\section{Conclusion}
\label{sec:conclusion}
We analyzed the impact of FPs and FNs against each other both in different regions and for two datasets with different architectures and center frequencies. Dense regions were found to be more resilient to detection errors and can tolerate moderate FP and FN rates, owing to their structural redundancy. Sparse regions, on the other hand, are highly sensitive to detection errors, with metrics dropping sharply as FP and FN rates increase. This underscores the need for stringent control over detection parameters in sparse regions to prevent rapid declines in PSNR and ensure high signal fidelity. The higher frequency introduced a slight increase in sensitivity to FN rates, but the overall impact remains less severe than in sparse regions.

Our findings further indicate that when using higher center frequencies, achieving optimal SSIM requires careful tuning of FP and FN rates, especially in sparse regions where misdetections have a more pronounced impact. The findings also underscore the need for adaptive detection thresholds that can account for regional density variations within the image. For applications requiring high spatial fidelity, such as dense microvascular imaging, it may be advantageous to employ higher frequencies while ensuring low FN rates through advanced detection algorithms. These insights could provide practical guidelines for clinicians and researchers to optimize ULM image quality by adjusting detection sensitivity based on MB density and imaging frequency.

\section{Acknowledgments}
\label{sec:acknowledgments}

This work was supported by the Natural Sciences and Engineering Research Council of Canada (NSERC) and the Government of Canada’s New Frontiers in Research Fund (NFRF), [NFRFE-2022-00295].

\bibliography{refs} 
\bibliographystyle{spiebib} 
\end{document}